%% file: main.tex
\documentclass[journal]{IEEEtran}

\pdfoutput=1
\usepackage{cite}

\ifCLASSINFOpdf
\else
\fi

\usepackage{amsmath}

\usepackage{algorithmic}
\usepackage{array}

\usepackage[english]{babel}
\usepackage[utf8]{inputenc}
\usepackage{amsmath}
\usepackage{graphicx}
\usepackage[colorinlistoftodos]{todonotes}
\usepackage{amsthm}
\usepackage{comment}
\usepackage{amssymb}
\usepackage{multirow}

\newcommand{\xref}{x_{ref}}
\usepackage{subcaption}
\usepackage{stringstrings}
\usepackage{xstring}
\usepackage{intcalc}
\usepackage{xparse}
\RequirePackage{expl3}
\usepackage{etoolbox}
\usepackage{pgfplots}
\usepackage[first=1, last=10]{lcg}
\usepackage{tikz}
\usetikzlibrary{calc}
\usetikzlibrary{shapes,arrows}
\usepackage{placeins}
\usepackage{curves}

\newcommand{\tb}[1]{\textbf{#1}}

\begin{document}
%
\title{Deep Local Binary Patterns}
%
%
%

\author{Kelwin~Fernandes
        ~and~Jaime~S.~Cardoso
\thanks{Kelwin Fernandes (kafc@inesctec.pt) and
Jaime S. Cardoso (jaime.s.cardoso@inesctec.pt) are with
\textit{Faculdade de Engenharia da Universidade do Porto} and
with INESC TEC, Porto, Portugal.}
}




\maketitle

\input{abstract}
\input{introduction}
\input{proposal}

\input{architectures}
\input{experiments}

\input{conclusions}

\section*{Acknowledgment}
This work was funded by the Project ``NanoSTIMA: Macro-to-Nano Human
Sensing: Towards Integrated Multimodal Health Monitoring and Analytics/-
NORTE-01-0145-FEDER-000016'' financed by the North Portugal Regional Op-erational Programme (NORTE 2020), under the PORTUGAL 2020 Partnership
Agreement, and through the European Regional Development Fund (ERDF),
and also by Funda\c{c}\~ao
para a Ci\^encia e a Tecnologia (FCT) within PhD grant
number SFRH/BD/93012/2013.

\bibliographystyle{IEEEtran}
\bibliography{references}

\end{document}

%% file: abstract.tex
\begin{abstract}
Local Binary Pattern (LBP) is a traditional descriptor
for texture analysis that gained attention in
the last decade. Being robust to several
properties such as invariance to illumination translation
and scaling, LBPs achieved state-of-the-art results in
several applications.
However, LBPs are not able to capture high-level features
from the image, merely encoding features with low abstraction
levels.
In this work, we propose Deep LBP,
which borrow ideas from the deep learning community to
improve LBP expressiveness. By using parametrized data-driven
LBP, we enable successive applications of the LBP operators with
increasing abstraction levels.
We validate the relevance
of the proposed idea in several datasets from a wide range
of applications. Deep LBP improved the performance of
traditional and multiscale LBP in all cases.
\end{abstract}

\begin{IEEEkeywords}
Local Binary Patterns, Deep Learning, Image Processing.
\end{IEEEkeywords}

%
\IEEEpeerreviewmaketitle

%% file: introduction.tex
\section{Introduction}
\label{sec:introduction}

%
%
%
%
\IEEEPARstart{I}{n} recent years, computer vision community has moved
towards the usage of deep learning strategies to solve a wide variety
of traditional problems, from image enhancement \cite{xie2012image} to scene
recognition \cite{zhou2014learning}.
Deep learning concepts emerged from traditional shallow concepts
from the early years of computer vision (e.g. filters, convolution, pooling,
thresholding, etc.).

Although these techniques have achieved \textit{state-of-the-art}
performance in several of tasks, the deep learning hype has
overshadowed research on other fundamental ideas.
Narrowing the spectrum of methods to a single class will eventually
saturate, creating a monotonous environment where the same basic idea
is being replicated over and over, and missing the opportunity to
develop other paradigms with the potential to lead to complementary
solutions. 

As deep strategies have benefited from traditional -shallow- methods in
the past, some classical methods started to take advantage of key deep
learning concepts. That is the case of deep Kernels \cite{cho2009kernel},
which  explores the successive application of nonlinear mappings within
the kernel umbrella.
In this work, we incorporate deep concepts into
Local Binary Patterns \cite{ojala2000gray,ojala2002multiresolution},
a traditional descriptor for texture analysis.
Local Binary Patterns (LBP) is a robust descriptor that briefly
summarizes texture information, being invariant to illumination
translation and scaling. LBP has been successfully used in a wide
variety of applications, including texture classification
\cite{liu2012extended,zhao2013completed,guo2010rotation,guo2010completed},
face/gender recognition
\cite{zhao2007dynamic,ahonen2004face,zhang2010local,ren2013noise,shan2012learning,%
huang2011local}, 
among others \cite{yeffet2009local,nanni2010local,xu2015adjustable}.

LBP has two main ingredients:

\begin{itemize}
\item The neighborhood (${\cal N}$), usually defined by an angular resolution
(typically 8 sampling angles) and radius $r$ of the neighborhoods.
Fig. \ref{fig:neighborhood} illustrates several possible neighborhoods.

\item The binarization function $b(\xref, x_i) \in \{0, 1\}$, which
allows the comparison between the reference point (central pixel) and
each one of the points $x_i$ in the neighborhood.
Classical LBP is applicable when $x_{ref}$ (and $x_{i}$) are in an ordered
set (e.g., $\mathbb{R}$ and $\mathbb{Z}$), with  $b(\xref, x_i)$ defined as

\begin{equation}
\label{eq:binfunction}
b(\xref, x_i) = (\xref \prec x_i), 
\end{equation}

where $\prec$ is the order relation on the set
(interpolation is used to compute $x_i$ when a neighbor location
does not coincide with the center of a pixel).

\end{itemize}

\input{figures/neighborhood}

The output of the LBP at each position $ref$ is the code resulting
from the comparison (binarization function) of the value $\xref$
with each of the $x_i$ in the neighborhood, with $i\in {\cal N}(ref)$,
see Figure \ref{fig:encoding}.
The LBP codes can be represented by their numerical value as 
formally defined in \eqref{eq:encoding}.

\begin{equation}
LBP(\xref) = \sum\limits_{i \in |{\cal N}(ref)|} 2^i \cdot b(\xref, x_i)
\label{eq:encoding}
\end{equation}

\input{figures/encoding}

LBP codes can take $2^{|\cal N|}$ different values.
In predictive tasks, for typical choices of angular resolution,
LBP codes are compactly summarized into a
histogram with $2^{|\cal N|}$ bins, being this the feature vector
representing the object/region/image (see Fig. \ref{fig:pipeline}).
Also, it is typical to compute the histograms in sub-regions and
to build a predictive model by using as features the concatenation
of the region histograms, being
non-overlapping and overlapping \cite{barkan2013fast} blocks traditional
choices (see Figure \ref{fig:multiblock}).

\input{figures/histograms}

In the last decade, several variations of the LBP have been proposed to attain
different properties. The two best-known variations were proposed by
Ojala et. al, the original authors of the LBP methodology: rotation invariant
and uniform LBP \cite{ojala2002multiresolution}.

Rotation invariance can be achieved by assigning a unique identifier to
all the patterns that can be obtained by applying circular shifts. The new
encoding is defined in \eqref{eq:riLBP}, where $ROR(e, i)$ applies a
circular-wrapped bit-wise shift of $i$ positions to the encoding $e$.

\begin{equation}
LBP(\xref) = \min \{ ROR(LBP(\xref), i)~|~i \in [0, \ldots |\cal N|) \}
\label{eq:riLBP}
\end{equation}

In the same work, Ojala et. al \cite{ojala2002multiresolution}
identified that uniform patterns (i.e. patterns with two or less circular
transitions) are responsible for the vast majority
of the histogram frequencies, leaving low discriminative power to the remaining
ones.
Then, the relatively small proportion of non-uniform patterns limits the reliability of their probabilities,
all the non-uniform patterns are assigned to a single bin in the
histogram construction, while uniform patterns are assigned to individual bins.
\cite{ojala2002multiresolution}.

Heikkila et. al proposed Center-Symmetric
LBP \cite{heikkila2006description}, which increases robustness on flat image areas 
and improves computational efficiency. This is achieved by comparing
pairs of neighbors located in centered symmetric directions
instead of comparing each neighbor with the reference pixel. Thus, an encoding
with four bits is generated from a neighborhood of 8 pixels. Also, the
binarization function incorporates an activation tolerance given by
$b(x_i, x_j) = x_i > x_j + T$.
Further extensions of this idea can be found in
\cite{trefny2010extended,xue2011hybrid,silva2015extended}.

Local ternary patterns (LTP) are an extension of LBP that use a 3-valued mapping
instead of a binarization function. The function used by LTP is formalized at
\eqref{eq:ltp}. LTP are less sensitive to noise in uniform regions at the cost
of losing invariance to illumination scaling. Moreover, LTP induce an additional
complexity in the number of encodings, producing histograms with up to
$3^{\cal N}$ bins.

\begin{equation}
b_T(\xref, x_i) = 
\begin{cases} 
	-1 & x_i < \xref - T \\
     0 & \xref -T \leq x_i \leq \xref + T\\
    +1 & \xref + T < x_i\\
\end{cases}
\label{eq:ltp}
\end{equation}

So far, we have mainly described methods that rely on redefining the construction
of the LBP encodings. A different line of
research focuses on improving LBP by modifying the texture summarization when
building the frequency histograms. Two examples of this idea were presented
in this work: uniform LBP \cite{ojala2002multiresolution} and multi-block LBP
(see Fig. \ref{fig:multiblock}). 

Since different scales may bring complementary information, one can
concatenate the histograms of LBP values at different scales. Berkan
et al. proposed this idea in the Over-Complete LBP (OCLBP)\cite{barkan2013fast}.
Besides computing the encoding at multiple scales, OCLBP computes the histograms
on several spatially overlapped blocks.
An alternative to this way of modeling multiscale patterns is to, at each point,
compute the LPB code at different scales, concatenate the codes and summarize
(i.e., compute the histogram) of the concatenated feature vector. 
This latter option has difficulties concerning the dimensionality of 
the data (potentially tackled with a bag of words approach) and the nature of 
the codes (making unsuitable standard k-means to find the bins-centers for 
a bag of words approach).

Multi-channel data (e.g. RGB) has been handled in a similar way, by 
1) computing and summarizing the LBP codes in each channel independently and then
concatenating the histograms \cite{choi2010using} and by 
2) computing a joint code for the three channels \cite{zhu2010multi}.

As LBP have been successfully used to describe spatial relations between pixels,
some works explored embedding temporal information on LBP for object detection 
and background removal
\cite{zhao2007dynamic,zhang2008dynamic,xue2010dynamic,yang2012spatio,%
yin2013dynamic,davarpanah2016texture}.

Finally, Local Binary Pattern Network (LBPNet) was introduced by
Xi et al. \cite{xi2016local}
as a preliminary attempt to embed Deep Learning concepts in
LBP. Their proposal consists on using a pyramidal approach on the
neighborhood scales and histogram sampling. Then, 
Principal Component Analysis (PCA) is used
on the frequency histograms to reduce the dimensionality of the feature
space.
Xi et al. analogise the pyramidal construction of LBP neighborhoods and
histogram sampling as a convolutional layer, where multiple filters
operate at different resolutions, and the dimensionality reduction
as a Pooling layer.
However, LBPNets aren't capable of aggregating information from a
single resolution into higher levels of abstraction which is the
main advantage of deep neural networks.

In the next sections, we will bring some ideas from the
Deep Learning community to traditional LBP. In this sense,
we intend to build LBP blocks that can be applied
recursively in order to build features with higher-level of
abstraction.

%% file: figures/neighborhood.tex
\newcommand{\lbpneighborhood}[5]{
\begin{subfigure}[t]{0.3\columnwidth}
\centering
\resizebox{\textwidth}{!}{
\begin{tikzpicture}

\def \grid {#1}
\def \halfgrid {#1 / 2}
\def \radius {#2}
\def \neighbors {#3}
\def \city {\equal{\unexpanded{#4}}{1}}
\pgfmathsetmacro\neighborsm{\neighbors - 1}

\draw[step=1cm,gray,very thin] (0,0) grid (\grid, \grid);

\draw[fill=gray!80,color=gray!30](\halfgrid, \halfgrid) circle (0.4cm);
\node at (\halfgrid, \halfgrid) {$\xref$};

\ifthenelse{\city}{
\draw[style=dashed,color=black!70](\halfgrid - \radius, \halfgrid - \radius)
	rectangle (\halfgrid + \radius, \halfgrid + \radius);
}{
\draw[style=dashed,color=black!70](\halfgrid, \halfgrid) circle (\radius cm);
}

\foreach \x in {0,...,\neighborsm}
	\ifthenelse{\city}{
    \draw [fill=black!80,color=black!70]
    		  ({\halfgrid + \radius * round(sin(\x*180*2/#3))},
           {\halfgrid + \radius * round(cos(\x*180*2/#3))}) circle (0.15cm);
	}
	{
    \draw [fill=black!80,color=black!70]
    		  ({\halfgrid + \radius * sin(\x*180*2/#3)},
           {\halfgrid + \radius * cos(\x*180*2/#3)}) circle (0.15cm);
	};
\end{tikzpicture}
}
\def \city {\equal{\unexpanded{#4}}{1}}
\caption{r=#2, n=#3\label{#5}}
\end{subfigure}
}

\begin{figure}[!t]
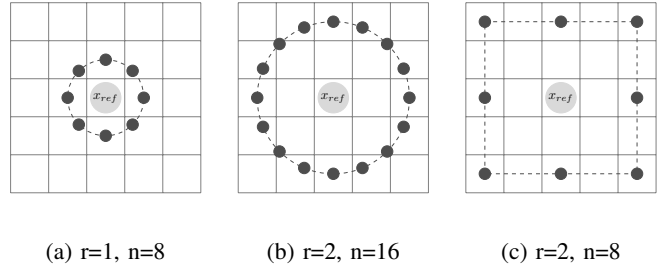

	\centering
	\lbpneighborhood{5}{1}{8}{0}{fig:neighborhood:10}
	\lbpneighborhood{5}{2}{16}{0}{fig:neighborhood:20}
	\lbpneighborhood{5}{2}{8}{1}{fig:neighborhood:11}
	\caption{LBP neighborhoods with radius ($r$) and angular resolution ($n$).
	The first two cases use Euclidean distance to define the neighborhood,
	the last case use Manhattan distance.
	\label{fig:neighborhood}}
\end{figure}

%% file: figures/encoding.tex
\ExplSyntaxOn

\newcounter{neighbors}
\newcounter{iterationid}

\newcommand{\lbpencoding}[3]{
\begin{subfigure}[t]{\columnwidth}
\centering
\resizebox{\textwidth}{!}{

\def\myresult{}

\setcounter{neighbors}{0}
\clist_map_inline:nn {#2}{
	\addtocounter{neighbors}{1}
}

\def\neighbors{\the\value{neighbors}}
\pgfmathsetmacro\neighborsm{\neighbors - 1}
\def\radius{0.25 * \columnwidth}
\def\angdist{360. / \neighbors}

\begin{tikzpicture}

\newcount\neighborid
\neighborid = 0

\draw[fill=black!30,color=black!30](\radius, \radius) circle (0.5cm);
\node at (\radius, \radius) {#1};

\setcounter{iterationid}{0}
\def\myresult{}

\draw[style=dashed,color=black!70](\radius, \radius) circle (\radius);

\clist_map_inline:nn {#2} {
	\def\sinshift{sin(\the\value{iterationid} * \angdist)}
	\def\cosshift{cos(\the\value{iterationid} * \angdist)}

	\draw[fill=black!70,color=black!70]
	({\radius - \radius * \sinshift}, {\radius + \radius * \cosshift})
	circle (0.4cm);

	\node[color=white] at 
	({\radius - \radius * \sinshift}, {\radius + \radius * \cosshift})
	{##1};

	\node at 
	({\radius - \radius * 0.65 * \sinshift},{\radius + \radius * 0.65 * \cosshift})
	{\ifthenelse{#1 < ##1}{$<$}{$>$}};


	\node at 
	({4 * \radius - \radius * \sinshift}, {\radius + \radius *\cosshift})
	{\ifthenelse{#1 < ##1}{1}{0}};
	
	\ifthenelse{#1 < ##1}{\xdef\myresult{1\myresult}}{\xdef\myresult{0\myresult}}

	\addtocounter{iterationid}{1}
};

\draw ({4 * \radius}, {\radius}) --
	  ({4 * \radius + 0.38 * 0.7 * \radius}, {\radius + 0.92 * 0.7 * \radius});
    
\node at (2.5*\radius, \radius) {\huge$\Rightarrow$};
\node at (5.5 * \radius, \radius) {\huge$\Rightarrow$};

\node at ({6.5 * \radius}, {\radius}){\huge\myresult};
\end{tikzpicture}
}

\caption{LBP~code~with~{\the\value{neighbors}}~neighbors.}
\end{subfigure}
}
\ExplSyntaxOff

\begin{figure}[!t]
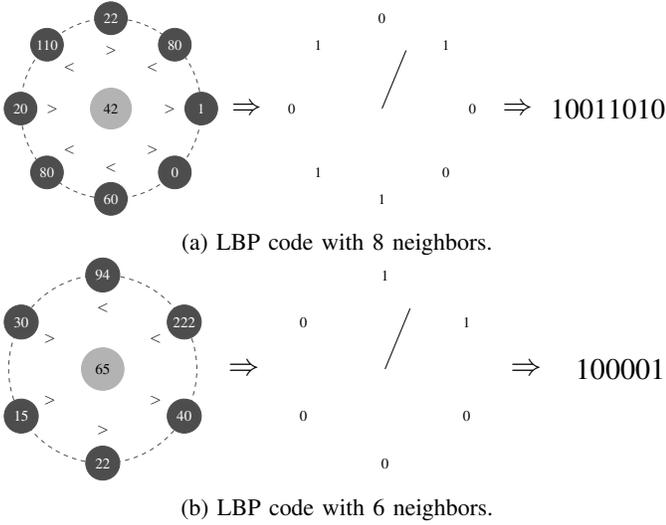

\lbpencoding{42}{22,110,20,80,60,0,1,80}
\\
\\
\lbpencoding{65}{94,30,15,22,40,222}{}

\caption{Cylinder and linear representation of the codes at some pixel positions.
Encodings are built in a clockwise manner from the starting point indicated in the
middle section of both figures.
\label{fig:encoding}}
\end{figure}

%% file: figures/histograms.tex
\pgfplotsset{
  compat=newest,
  xlabel near ticks,
  ylabel near ticks
}

\newcommand{\random}{\rand\arabic{rand}}

\newcommand{\encodingmatrix}[2]{
\resizebox{\columnwidth}{!}{

\begin{tikzpicture}
	\def\hsize{#1};
	\def\prevhsize{#2};
	\draw[step=1cm,gray,very thin] (0,0) grid (\hsize, \hsize);
	\foreach \r in {1,...,\hsize}
	{
		\def\emptyrow{\r=\prevhsize}
		\def\emptycol{\c=\prevhsize}
		\foreach \c in {1,...,\hsize}%
		{
			\ifthenelse{\emptycol \AND \emptyrow}
			{\node at (\c - 0.5, \hsize - \r + 0.5) {$\ddots$};}
			{}
			\ifthenelse{\emptyrow \AND \NOT{\emptycol}}
			{\node at (\c - 0.5, \hsize - \r + 0.5) {$\vdots$};}
			{}%
			\ifthenelse{\NOT{\emptyrow} \AND \emptycol}
			{\node at (\c - 0.5, \hsize - \r + 0.5) {$\hdots$};}
			{}%
			\ifthenelse{\NOT{\emptyrow} \AND \NOT{\emptycol}}
			{\node at (\c - 0.5, \hsize - \r + 0.5) {$e_{\random}$}}
			{};
		}%
	};
\end{tikzpicture}
}
}

\newcommand{\buildhist}[5]{
\begin{tikzpicture}
	\pgfplotsset{every tick label/.append style={font=\huge}}
    \begin{axis}[
      ybar,
      bar width=20pt,
      xlabel={Encodings},
      ylabel={Frequency},
      ytick=\empty,
      xtick=data,
      axis x line=bottom,
      axis y line=left,
      enlarge x limits=0.2,
      symbolic x coords={a, b, c, d, e},
      xticklabel style={anchor=base,yshift=-\baselineskip},
      xticklabels={$e_1$, $e_2$, $e_3$, $\ldots$, $e_n$},
      xticklabel style = {yshift=-0.4cm},
    ]
    \addplot[fill=#5] coordinates {
        (a,#1)
        (b,#2)
        (c,#3)
        (d,0)
        (e,#4)
      };
    \end{axis}
\end{tikzpicture}%
}

\begin{figure}[!t]
	\resizebox{\columnwidth}{!}{
		\begin{tabular}{ccccccc}
			\Huge{Original image} &
			\hspace*{2cm}\Huge{LBP encodings} & &
			\Huge{Histogram} & &
			\\
			\\
			\begin{minipage}[t]{0.46\textwidth}
			\centering
			\includegraphics[width=\textwidth]{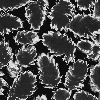}
			\end{minipage}
			&
			\begin{tikzpicture}
				\draw[fill=white,color=white]
				(0,0) -- (2,0) -- (2,8) -- (0,8) -- (0,0);
				\node at (1, 4) {\Huge{$\Rightarrow$}};
			\end{tikzpicture}
			\encodingmatrix{5}{4} &
			\begin{tikzpicture}
				\draw[fill=white,color=white]
				(0,0) -- (2,0) -- (2,8) -- (0,8) -- (0,0);
				\node at (1, 4) {\Huge{$\Rightarrow$}};
			\end{tikzpicture} & 
			\buildhist{10}{30}{10}{60}{white} &
			\begin{tikzpicture}
				\draw[fill=white,color=white]
				(0,0) -- (2,0) -- (2,8) -- (0,8) -- (0,0);
				\node at (1, 4) {\Huge{$\Rightarrow$}};
			\end{tikzpicture} & 
			\begin{tikzpicture}
				\draw[fill=gray!30,color=gray!30]
				(0,0) -- (4,0) -- (4,8) -- (0,8) -- (0,0);
				\node at (2, 4) {\Huge{Classifier}};
			\end{tikzpicture}
		\end{tabular}
	}

\caption{Traditional pipeline for image classification using LBP.
\label{fig:pipeline}}
\end{figure}
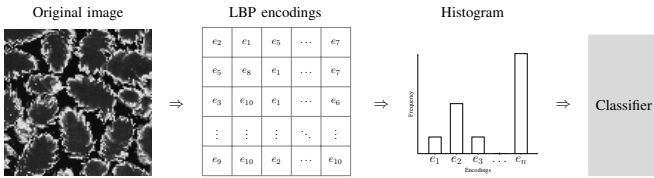


\begin{figure}[!t]
\resizebox{0.8\columnwidth}{!}{
	\begin{minipage}{\textwidth}
		\resizebox{\columnwidth}{!}{
		\begin{tikzpicture}
			\def\hsize{8};
			\def\prevhsize{7};
			\draw[fill=blue!30,color=gray!10]
				(0,0) -- (4,0) -- (4,4) -- (0,4) -- (0,0);
			\draw[fill=red!30,color=gray!27]
				(4,0) -- (8,0) -- (8,4) -- (4,4) -- (4,0);
			\draw[fill=green!30,color=gray!54]
				(4,4) -- (8,4) -- (8,8) -- (4,8) -- (4,4);
			\draw[fill=yellow!30,color=gray!90]
				(0,4) -- (4,4) -- (4,8) -- (0,8) -- (0,4);
			\draw[step=1cm,gray,very thin,color=black] (0,0) grid (\hsize, \hsize);
		\end{tikzpicture}
		}
	\end{minipage}
	\begin{minipage}{0.6\textwidth}
		\begin{center}
		\buildhist{10}{20}{70}{50}{gray!90}\\
		\buildhist{10}{30}{10}{60}{gray!54}\\
		\buildhist{25}{50}{90}{5}{gray!27}\\
		\buildhist{90}{30}{80}{10}{gray!10}\\	
		\end{center}
	\end{minipage}
	\begin{minipage}{0.1\columnwidth}
		\begin{flushleft}
		\begin{tikzpicture}
			\def\hsize{8};
			\def\prevhsize{7};
			\draw (0,20) -- (2,20);
			\draw (0,0) -- (2,0);
			\draw (0,6.66) -- (2,6.66);
			\draw (0,13.33) -- (2,13.33);
			\draw (2,10) -- (4,10);
			\draw (0,0) -- (2,0);
			\draw (2,0) -- (2,20);
			\draw[fill=gray!30,color=gray!30]
				(4,6.66) -- (8,6.66) -- (8,13.33) -- (4,13.33) -- (4,6.66);
			\node at (6, 10) {\Huge{Classifier}};
		\end{tikzpicture}
		\end{flushleft}
	\end{minipage}
}

\caption{Multi-block LBP with $2\times 2$ non-overlapping blocks.
\label{fig:multiblock}}
\end{figure}
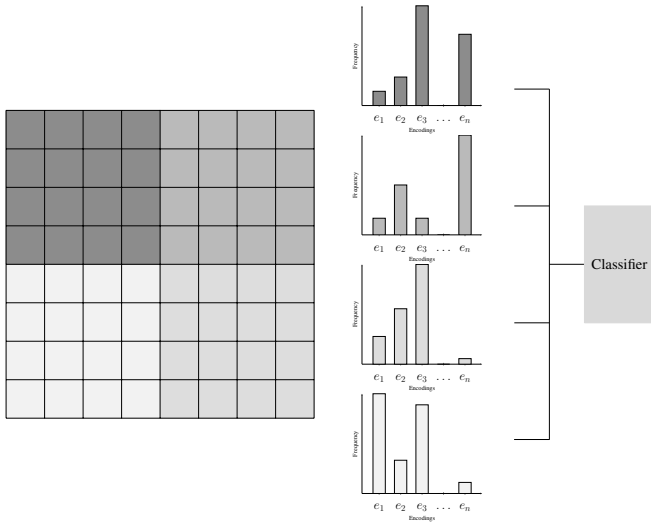

%% file: proposal.tex
\section{Deep Local Binary Patterns}
\label{sec:proposal}

The ability to build features with increasing abstraction level using a
recursive combination of relatively simple processing modules is one of the
reasons that made Convolutional Neural Networks -- and in general Neural
Networks -- so successful. In this work, 
we propose to represent ``higher order'' information about texture by applying
LBP recursively, i.e., cascading LBP computing blocks one after the other
(see Fig \ref{fig:deepLBP}). In this sense, while an LBP encoding describes
the local texture, a second order LBP encoding describes the texture
of textures.

\input{figures/recursivelbp}

However, while it is trivial to recursively apply convolutions -- and many other
filters -- in a cascade fashion,
traditional LBP are not able to \textit{digest} their own output.

Traditional LBP rely on receiving as input an image with domain
in an ordered set (e.g. grayscale intensities).
However, LPB codes are not in an ordered set, dismissing the direct recursive
application of standard LBP.
As such we will first generalize the main operations supporting LBP and
discuss next how to assemble together a deep/recursive LBP feature
extractor. We start the discussion with the binarization function
$b(\xref, x_i)$. 

It is instructive to think, in the conventional LBP, if non-trivial alternative functions exist to the adopted one, Eq. \eqref{eq:binfunction}. What is(are) the main property(ies) required by the binarization function? Does it need to make use of a (potentially implicit) order relationship?
A main property of the binarization function is to be independent of scaling and translation of the input data, that is,

\begin{equation}
b(k_1 \xref + k_2,~k_1 x_{i} + k_2) = b(\xref, x_i),~k_1 > 0.\label{eq:binproperties}
\end{equation}

It is trivial to prove that the only options for the binarization function that
hold Eq. \eqref{eq:binproperties} are the constant functions
(always zeros or always ones), the one given by Eq. \eqref{eq:binfunction}
and its reciprocal.

\begin{proof}
Assume $x_i,x_j > \xref$ and $b(\xref,x_i) \not = b(\xref,x_j)$.
Under the independence to translation and scaling (Eq. \eqref{eq:binproperties}),
$b(x_{ref},x_i) = b(x_{ref},x_j)$ as shown below,
which is a contradiction.

\def\kt{\dfrac{x_i-\xref}{x_j - \xref}}
\def\kc{\xref\dfrac{x_j-x_i}{x_j - \xref}}

\resizebox{\columnwidth}{!}{
\begin{minipage}{\linewidth}
\begin{align*}
  & b(\xref, x_i)\\
= & \left\langle~\text{Identity of multiplication}~\right\rangle\\
  & b\left(\dfrac{x_j - \xref}{x_j - \xref}\xref,
  		   \dfrac{x_j - \xref}{x_j - \xref}x_i 
  	 \right)\\
= & \left\langle~\text{Identity of addition}~\right\rangle\\
  & b\left(\dfrac{x_j - \xref + (x_i - x_i)}{x_j - \xref}\xref,
  		   \dfrac{x_j - \xref}{x_j - \xref}x_i
  		   + \left(\dfrac{\xref x_j - \xref x_j}{x_j - \xref}\right)
  	 \right)\\
= & \left\langle \text{Arithmetic} \right\rangle\\
  & b\left(\kt \xref + \kc, \kt x_j + \kc\right)\\
= & \left\langle~\text{Eq.}~\eqref{eq:binproperties}\text{, where}~
	k_1 = \kt, k_2 = \kc
	\right\rangle\\
  & b(\xref, x_j)\\
\end{align*}
\end{minipage}
}

Therefore, $b(\xref, x_i)$ must be equal to all $x_i$ above $\xref$.
Similarly, $b(\xref, x_i)$ must be equal to all $x_i$ below $\xref$.
\end{proof}

Among our options, the constant binarization function is not a viable option,
since the information (in information theory perspective) in the output is zero. 
Since the recursive application of functions can be understood as a composition,
invariance to scaling and translation is trivially ensured by using a
traditional LBP in the first transformation.

Given that transitivity is a relevant property held by
the natural ordering of real numbers, we argue that
such property should be guaranteed by our
binarization function. In this sense, we will focus
on strict partial orders of encodings.
Following, we show how to build such binarization functions for
the i-$th$ application of the LBP operator, where $i > 1$.
We will consider both predefined/expert driven solutions and
data driven solutions (and therefore, application specific).

Hereafter, we will refer to the binarization function as the 
partial ordering of LBP encodings. Despite the existence of other
types of functions may be of general interest, narrowing the
search space to those that can be represented as a partial
ordering induce efficient learning mechanisms.

\subsection{Preliminaries}

\def\EN{E_{\cal N}}
\def\db{b^+}

Let us formalize the deep binarization function as the order relation
$\db \in {\cal P}(\EN \times \EN)$, where $\EN$ is the set of encodings
induced by the neighborhood $\cal N$.

Let $\Phi$ be an oracle
$\Phi :: {\cal P}(\EN \times \EN) \rightarrow \mathbb{R}$ that assesses
the performance of a binarization function. For example, 
among other options,
the oracle can be defined as the performance of the
traditional LBP pipeline (see Fig. \ref{fig:pipeline}) on a dataset
for a given predictive task.

\subsection{Deep Binarization Function}

From the entire space of binarization functions, we restrict our analysis
to those induced by strict partial orders. Within this context,
it is easy to see that learning the best binarization function by
exhaustive exploration is intractable since the number of combinations
equals the number of directed acyclic graph (DAG) with
$2^{|\cal N|} = |E_{\cal N}|$ nodes.
The DAG counting problem was studied by Robinson \cite{robinson1977counting}
and is given by the recurrence relation Eqs.
\eqref{eq:dagcounting1}-\eqref{eq:dagcounting2}.

\begin{align}
a_0 & = 1\label{eq:dagcounting1}\\
a_{n>1} & = \sum\limits_{k=1}^n (-1)^{k - 1} \binom{n}{k}2^{k(n - k)}a_{n - k}
\label{eq:dagcounting2}
\end{align}

\input{tables/nencodings}

Table \ref{tab:nenc} illustrates the size of the search space for
several numbers of neighbors. For instance, for the traditional
setting with 8 neighbors, the number of combinations has more
than 10,000 digits. Thereby, a heuristic approximation must
be carried out.

\subsubsection{Learning $\db$ from a User-defined dissimilarity function}

The definition of a dissimilarity function between two codes seems reasonably
accessible. For instance, an immediate option is to adopt the hamming
distance between codes, $d_H$. With rotation invariance in mind, one can
adopt the minimum hamming distance between all circularly shifted
versions of $x_{ref}$ and $x_i$, $d_H^{ri}$.
The circular invariant hamming distance between $\xref$ and $x_i$ can be 
computed as

\begin{equation}
d_H^{ri} = \min\limits_{s \in {0, \cdots, {\cal N}-1}} d_H (ROR(\xref, s), x_i)
\end{equation}

Having defined such a dissimilarity function between pairs of codes, one can
know proceed with the definition of the binarization function.

Given the dissimilarity function, we can learn a mapping of the codes
to an ordered set. Resorting to Spectral Embedding \cite{ng2001spectral}, one can obtain such a mapping.
The conventional binarization function,
Eq.~\eqref{eq:binfunction},
can then be applied. Other alternatives for building the mappings
can be found in the manifold learning literature: Isomaps \cite{tenenbaum2000global},
Multidimensional scaling (MDS) \cite{kruskal1964nonmetric}, among others.
In this case, the oracle function can be defined as the
intrinsic loss functions used in the optimization process of
such algorithms.

Preserving a desired property $P$ such as rotational invariance and sign invariance
(i.e. interchangeability between ones and zeros) can be achieved by considering
$P$-aware dissimilarities.

\subsubsection{Learning $\db$ from a High-dimensional Space}
\label{sec:proposal:high}

A second option is to map the code space to a new (higher-dimensional) space that
characterizes LBP encodings. Then, an ordering or preference relationship can be
learned in the extended space, for instance resorting to preference learning
algorithms \cite{fernandes2016learning,joachims2002optimizing,flach2007simple}.

Some examples of properties that characterize LBP encodings are:

\begin{itemize}
	\item Number of transitions of size 1 (e.g. 101, 010).
	\item Number of groups/transitions.
	\item Size of the smallest/largest group.
	\item Diversity on the group sizes.
	\item Number of ones.
\end{itemize}

Techniques to learn the final ordering based on the new high-dimensional
space include:

\begin{itemize}
	\item Dimensionality reduction techniques, including Spectral embeddings,
	PCA, MDS and other manifold techniques.
	\item Preference learning strategies for learning rankings
	\cite{fernandes2016learning,joachims2002optimizing,flach2007simple}.
\end{itemize}

A case of special interest that will be used in the experimental section of
this work are Lexicographic Rankers (LR) \cite{flach2007simple,fernandes2016learning}.
In this work, we will focus on the simplest type of LR, linear LR.
Let us assume that features in the new high-dimensional space are
scoring rankers (e.g. monotonic functions) on the texture complexity of the codes.
Thus, for each code $e_i$ and feature $s_j$, the complexity associated to $e_i$
by $s_j$ is denoted as $s_j(e_i)$. We assume $s_j(e_i)$ to lie in a discrete
domain with a well-known order relation.

Thus, each feature is grouping the codes into equivalence classes. For example,
the codes with 0 transitions (i.e. flat textures), 2 transitions
(i.e. uniform textures) and so on.

If we concatenate the output of the scoring rankers in a linear manner
$(s_0(e_i), s_1(e_i), \cdots, s_n(e_i))$, a natural arrangement is
their lexicographic order (see Eq. \eqref{eq:lexord}), where each
$s_j(e_i)$ is subordering the equivalence class obtained by the
previous prefix of rankers
$(s_0(e_i), \cdots, s_{j-1}(e_i))$.

\begin{equation}
\text{LexRank}(a, b) =
\begin{cases}
a = b &, |a| = 0 ~\vee |b| = 0\\
a \prec b &, a_0 \prec b_0\\
a \succ b &, a_0 \succ b_0\\
LexRank(t(a), t(b)) &, a_0 = b_0\\
\end{cases}
\label{eq:lexord}
\end{equation}

\noindent
where $t(a)$ returns the tail of the sequence. Namely, the order
between two encodings is decided by the first scoring ranker in
the hierarchy that assigns different values to the encodings.

Therefore, the learning process is reduced to find the best feature arrangement.
A heuristic approximation to this problem can be achieved by iteratively 
appending to the sequence of features the one that maximizes the performance
of the oracle $\Phi$.

Similarly to property-aware dissimilarity functions, 
if the features in the new feature vector ${\cal V} (x)$ are invariant
to $P$, the $P$-invariance of the learned binarization function
is automatically guaranteed.

%% file: figures/recursivelbp.tex
\tikzstyle{block} = [rectangle, draw, text centered, rounded corners,
					 minimum height=5em, minimum width=2em]
\tikzstyle{line} = [draw, -latex']
\tikzstyle{cloud} = [rectangle, draw, text centered, rounded corners,
					 minimum height=5em, minimum width=2em, fill=gray!20]

\begin{figure}[!t]
\centering
\begin{tikzpicture}[node distance = 1.2cm, auto]
    \node [cloud] (img) {\rotatebox{90}{Image}};
    \node [block, right of=img] (lbp0) {\rotatebox{90}{LBP}};
    \node [block, right of=lbp0] (lbp1) {\rotatebox{90}{LBP}};
    \node [right of=lbp1] (dots) {$\hdots$};    
    \node [block, right of=dots] (lbpn) {\rotatebox{90}{LBP}};
    \node [cloud, right of=lbpn] (hist) {\rotatebox{90}{Histogram}};
    \node [block, right of=hist] (model) {\rotatebox{90}{Model}};

    \path [line] (img) -- (lbp0);
    \path [line] (lbp0) -- (lbp1);
    \path [line] (lbp1) -- (dots);
    \path [line] (dots) -- (lbpn);
    \path [line] (lbpn) -- (hist);
    \path [line] (hist) -- (model);
\end{tikzpicture}

\caption{Recursive application of Local Binary Patterns.\label{fig:deepLBP}}
\end{figure}
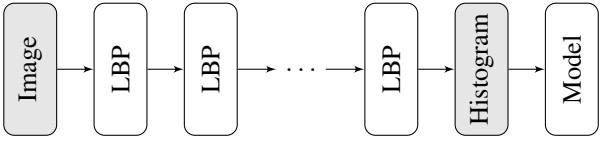

%% file: tables/nencodings.tex
\begin{table}[!t]
\caption{\label{tab:nenc}Lower bound of the number of combinations
for deciding the best LBP binarization function as \textit{partial orders}.}
\centering
\begin{tabular}{|c|ccc|}
\hline
\textbf{\# Neighbors} & \textbf{Rotational Inv.} & \textbf{Uniform} & \textbf{Traditional}\\
\hline
$  2$ & $2 \cdot 10^{    1}$ & $2 \cdot 10^{    4}$ & $5 \cdot 10^{    2}$\\
$  3$ & $5 \cdot 10^{    2}$ & $1 \cdot 10^{   15}$ & $7 \cdot 10^{   11}$\\
$  4$ & $3 \cdot 10^{    6}$ & $2 \cdot 10^{   41}$ & $8 \cdot 10^{   46}$\\
$  5$ & $7 \cdot 10^{   11}$ & $6 \cdot 10^{   94}$ & $2 \cdot 10^{  179}$\\
$  6$ & $1 \cdot 10^{   36}$ & $2 \cdot 10^{  190}$ & $1 \cdot 10^{  685}$\\
$  7$ & $2 \cdot 10^{   72}$ & $3 \cdot 10^{  346}$ & $3 \cdot 10^{ 2640}$\\
$  8$ & $1 \cdot 10^{  225}$ & $1 \cdot 10^{  585}$ & $3 \cdot 10^{10288}$\\
\hline
\end{tabular}
\end{table}

%% file: architectures.tex
\section{Deep Architectures}

Given the closed form of the LBP with deep binarization
functions, their recursive combination seems feasible.
In this section, several alternatives for the aggregation
of deep LBP operators are proposed.

\subsection{Deep LBP (DLBP)}

\input{figures/architectures/deeplbp}

The simplest way of aggregating Deep LBP operators is by
applying them recursively and computing the final encoding
histograms. Figure \ref{fig:arch:deepLBP} shows this
architecture. The first transformation is done by a traditional
shallow LBP while the remaining transformations are performed
using deep binarization functions.

Figure \ref{fig:deepvis} illustrates the patterns detected
by several deep levels on a cracker image from the Brodatz
database.
In this case, the ordering between LBP encodings was learned
by using a lexicographic ordering of encodings on the
number of groups, the size of the largest group and imbalance
ratio between 0's and 1's.
We can observe that the initial layers of the
architecture extract information from local textures while the
later layers have higher levels of abstraction.

\input{figures/deepvisualization}

\subsection{Multi-Deep LBP (MDLBP) \label{sec:architectures:multideeplbp}
}

Despite it may be a good idea to extract higher-order
information from images, for the usual applications of LBP,
it is important to be able to detect features at different
levels of abstraction. For instance, if the image has
textures with several complexity levels, it may be relevant
to keep the patterns extracted at different abstraction
levels. Resorting to the techniques employed in the analysis
of multimodal data \cite{kapoor2005multimodal},
we can aggregate this information
in two ways: feature and decision-level fusion.

\subsubsection{Feature-level fusion} one histogram is computed
at each layer and the model is built using the concatenation
of all the histograms as features.
\subsubsection{Decision-level fusion} one histogram and decision
model is computed at each layer. The final model uses the
probabilities estimated by each individual model to produce
the final decision.

Figures \ref{fig:arch:deepLBPearly} and
\ref{fig:arch:deepLBPlate} show Multi-Deep LBP architectures
with feature-level and decision-level fusion respectively.
In our experimental setting, feature-level fusion was used.

\input{figures/architectures/architectures}

\subsection{Multiscale Deep LBP (Multiscale DLBP)}

\input{figures/architectures/multiscaledeeplbp}

In the last few years, deep learning approaches have benefited
from multi-scale architectures that are able to aggregate information
from different image scales
\cite{eigen2014depth, neverova2014multi, wang2014learning}. Despite
being able to induce higher abstraction levels, deep networks are
restrained to the size of the individual operators. Thereby, 
aggregating multi-scale information in deep architectures may
exploit their capability to detect traits that appear at different
scales in the images in addition to turning the decision process
scale invariant.

In this work, we consider the stacking of independent deep architectures
at several scales. The final decision is done by concatenating the
individual information produced at each scale factor (cf. Figure
\ref{fig:arch:multiscaledeepLBP}). Depending on the fusion approach,
the final model operates in different spaces
(i.e. feature or decision level).
In an LBP context, we can define the scale operator of an image
by resizing the image or by increasing the neighborhood radius.

%% file: figures/architectures/deeplbp.tex
\tikzstyle{block} = [rectangle, draw, text centered, rounded corners,
					 minimum height=5em, minimum width=2em]
\tikzstyle{line} = [draw, -latex']
\tikzstyle{cloud} = [rectangle, draw, text centered, rounded corners,
					 minimum height=5em, minimum width=2em, fill=gray!20]

\tikzstyle{lbpnode} = [rectangle, draw, text centered, rounded corners,
				      minimum height=2em, minimum width=5em]
\tikzstyle{proposed} = [lbpnode]
\tikzstyle{line} = [draw, -latex']
\tikzstyle{traditional} = [lbpnode, fill=gray!30]
\tikzstyle{image} = [rectangle, draw, text centered,
    		   			 minimum width=4em, minimum height=4em, fill=gray!80]

\begin{figure}[!t]
\centering
\resizebox{\columnwidth}{!}{
\begin{tikzpicture}[node distance = 2cm, auto]
    \node [image] (img) {Image};
    \node [traditional, right of=img] (lbp0) {LBP};
    \node [proposed, right of=lbp0] (lbp1) {LBP-b$_1^+$};
    \node [right of=lbp1, node distance=1.5cm] (dots) {$\cdots$};    
    \node [proposed, right of=dots, node distance=1.5cm] (lbpn)
    {LBP-b$_n^+$};
    \node [traditional, below of=lbpn, node distance=1cm] (hist) {Histogram};
    \node [traditional, below of=hist, node distance=1cm] (model) {Model};

    \path [line] (img) -- (lbp0);
    \path [line] (lbp0) -- (lbp1);
    \path [line] (lbp1) -- (dots);
    \path [line] (dots) -- (lbpn);
    \path [line] (lbpn) -- (hist);
    \path [line] (hist) -- (model);
\end{tikzpicture}
}
\caption{Deep LBP.\label{fig:arch:deepLBP}}
\end{figure}
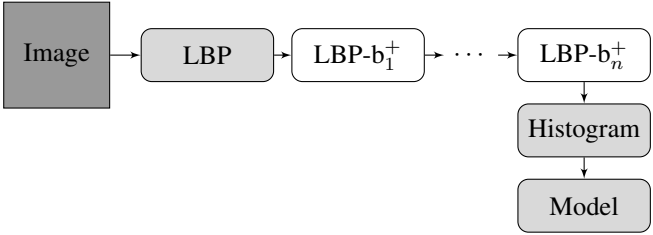

%% file: figures/deepvisualization.tex
\def\imgsize{0.18\textwidth}

\begin{figure}[!t]
	\centering
	\begin{subfigure}[t]{\imgsize}
        \centering
        \includegraphics[width=\textwidth]
			{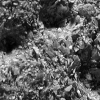}
        \caption{Input image}
	\end{subfigure}
	\begin{subfigure}[t]{\imgsize}
        \centering
        \includegraphics[width=\textwidth]
	        {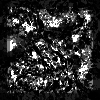}
        \caption{LBP}
	\end{subfigure}

	\begin{subfigure}[t]{\imgsize}
        \centering
        \includegraphics[width=\textwidth]
			{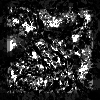}
        \caption{DeepLBP(1)}
	\end{subfigure}
	\begin{subfigure}[t]{\imgsize}
        \centering
        \includegraphics[width=\textwidth]
			{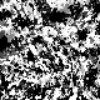}
        \caption{DeepLBP(2)}
	\end{subfigure}

	\begin{subfigure}[t]{\imgsize}
        \centering
        \includegraphics[width=\textwidth]
	        {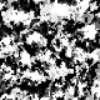}
        \caption{DeepLBP(3)}
	\end{subfigure}
	\begin{subfigure}[t]{\imgsize}
        \centering
        \includegraphics[width=\textwidth]
	        {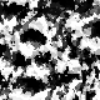}
        \caption{DeepLBP(4)}
	\end{subfigure}

	\caption{Visualization of LBP encodings from a
	Brodatz database \cite{brodatz1966textures} image.
	The results obtained by applying n layers
	of Deep LBP operators are denoted as DeepLBP(n).
	A neighborhood of size 8, radius 10 and Euclidean
	distance was used.
	The grayscale intensity is defined by the order of the
	equivalence classes.
	\label{fig:deepvis}}
\end{figure}

%% file: figures/architectures/architectures.tex
\tikzstyle{block} = [rectangle, draw, text centered, rounded corners,
					 minimum height=5em, minimum width=2em]
\tikzstyle{line} = [draw, -latex']
\tikzstyle{cloud} = [rectangle, draw, text centered, rounded corners,
					 minimum height=5em, minimum width=2em, fill=gray!20]

\begin{figure}[!t]
	\centering
	\input{figures/architectures/deeplbpearly}
	\input{figures/architectures/deeplbplate}

	\caption{Deep LBP architectures.\label{fig:archictectures}}
\end{figure}
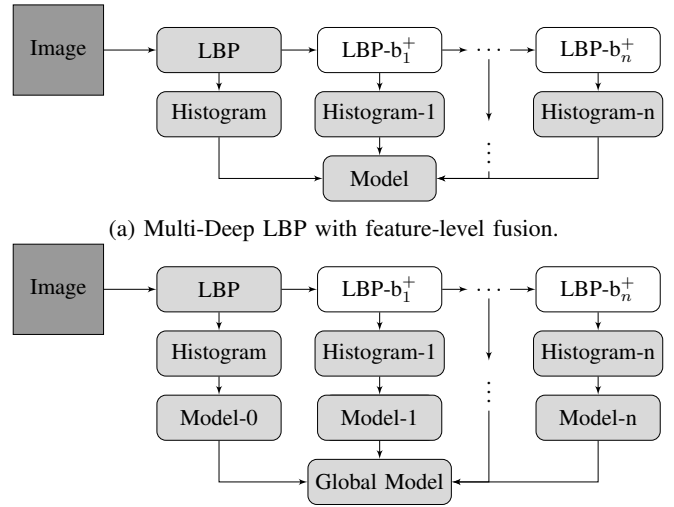

%% file: figures/architectures/deeplbpearly.tex
\tikzstyle{lbpnode} = [rectangle, draw, text centered, rounded corners,
				      minimum height=2em, minimum width=5.5em]
\tikzstyle{proposed} = [lbpnode]
\tikzstyle{line} = [draw, -latex']
\tikzstyle{traditional} = [lbpnode, fill=gray!30]
\tikzstyle{image} = [rectangle, draw, text centered,
    		   			 minimum width=4em, minimum height=4em, fill=gray!80]

\begin{subfigure}[t]{\columnwidth}
\centering
\resizebox{\columnwidth}{!}{
\begin{tikzpicture}[node distance = 2.5cm, auto]
    \node [image] (img) {Image};
    \node [traditional, right of=img] (lbp0) {LBP};
    \node [proposed, right of=lbp0] (lbp1) {LBP-b$_1^+$};
    \node [right of=lbp1, node distance=1.7cm] (dots) {$\hdots$};    
    \node [proposed, right of=dots, node distance=1.7cm] (lbpn) {LBP-b$_n^+$};

    \node [traditional, below of=lbp0,
    		   node distance=1cm] (hist0) {Histogram};
    \node [traditional, below of=lbp1,
    		   node distance=1cm] (hist1) {Histogram-1};
    \node [below of=dots, node distance=1.5cm] (vdots) {$\vdots$};    
    \node [traditional, below of=lbpn,
    		   node distance=1cm] (histn) {Histogram-n};

    \node [traditional, below of=hist1,
    		   node distance=1cm, minimum height=2em,
    		   minimum width=5em] (model) {Model};

    \path [line] (img) -- (lbp0);
    \path [line] (lbp0) -- (lbp1);
    \path [line] (lbp1) -- (dots);
    \path [line] (dots) -- (lbpn);

    \path [line] (lbp0) -- (hist0);
    \path [line] (lbp1) -- (hist1);
    \path [line] (dots) -- (vdots);
    \path [line] (lbpn) -- (histn);

    \path [line] (hist0) |- (model);
    \path [line] (hist1) -- (model);
    \path [line] (vdots) |- (model);
    \path [line] (histn) |- (model);
\end{tikzpicture}
}
\caption{Multi-Deep LBP with feature-level fusion.
\label{fig:arch:deepLBPearly}}
\end{subfigure}

%% file: figures/architectures/deeplbplate.tex
\tikzstyle{lbpnode} = [rectangle, draw, text centered, rounded corners,
				      minimum height=2em, minimum width=5.5em]
\tikzstyle{proposed} = [lbpnode]
\tikzstyle{line} = [draw, -latex']
\tikzstyle{traditional} = [lbpnode, fill=gray!30]
\tikzstyle{image} = [rectangle, draw, text centered,
    		   			 minimum width=4em, minimum height=4em, fill=gray!80]

\begin{subfigure}[t]{\columnwidth}
\centering
\resizebox{\columnwidth}{!}{
\begin{tikzpicture}[node distance = 2.5cm, auto]
    \node [image] (img) {Image};
    \node [traditional, right of=img] (lbp0) {LBP};
    \node [proposed, right of=lbp0] (lbp1) {LBP-b$_1^+$};
    \node [right of=lbp1, node distance=1.7cm] (dots) {$\hdots$};    
    \node [proposed, right of=dots, node distance=1.7cm] (lbpn) {LBP-b$_n^+$};

    \node [traditional, below of=lbp0,
    		   node distance=1cm] (hist0) {Histogram};
    \node [traditional, below of=lbp1,
    		   node distance=1cm] (hist1) {Histogram-1};
    \node [below of=dots, node distance=1.5cm] (vdots) {$\vdots$};    
    \node [traditional, below of=lbpn,
    		   node distance=1cm] (histn) {Histogram-n};

    \node [traditional, below of=hist1,
    		   node distance=1cm, minimum height=2em,
    		   minimum width=5em] (model) {Model};

    \node [traditional, below of=hist0,
    		   node distance=1cm] (m0) {Model-0};
    \node [traditional, below of=hist1,
    		   node distance=1cm] (m1) {Model-1};
    \node [traditional, below of=histn,
    		   node distance=1cm] (mn) {Model-n};

    \node [traditional, below of=m1,
    		   node distance=1cm, minimum height=2em,
    		   minimum width=5em] (model) {Global Model};

    \path [line] (img) -- (lbp0);
    \path [line] (lbp0) -- (lbp1);
    \path [line] (lbp1) -- (dots);
    \path [line] (dots) -- (lbpn);

    \path [line] (lbp0) -- (hist0);
    \path [line] (lbp1) -- (hist1);
    \path [line] (dots) -- (vdots);
    \path [line] (lbpn) -- (histn);

    \path [line] (hist0) -- (m0);
    \path [line] (hist1) -- (m1);
    \path [line] (vdots) |- (model);
    \path [line] (histn) -- (mn);

    \path [line] (m0) |- (model);
    \path [line] (m1) -- (model);
    \path [line] (mn) |- (model);
\end{tikzpicture}
}
\caption{Multi-Deep LBP with decision-level fusion.
\label{fig:arch:deepLBPlate}}
\end{subfigure}

%% file: figures/architectures/multiscaledeeplbp.tex
\tikzstyle{lbpnode} = [rectangle, draw, text centered, rounded corners,
				       minimum height=2em, minimum width=5.5em]
\tikzstyle{proposed} = [lbpnode]
\tikzstyle{line} = [draw, -latex']
\tikzstyle{traditional} = [lbpnode, fill=gray!30]
\tikzstyle{histogramblock} = [traditional,node distance=1.5cm]
\tikzstyle{image} = [rectangle, draw, text centered,
    		   			 minimum width=6em, minimum height=6em, fill=gray!80]
\tikzstyle{deepbox} = [rectangle, draw, text centered,
    		   			   minimum width=6em, minimum height=3em, fill=black!90,
    		   			   text=white]

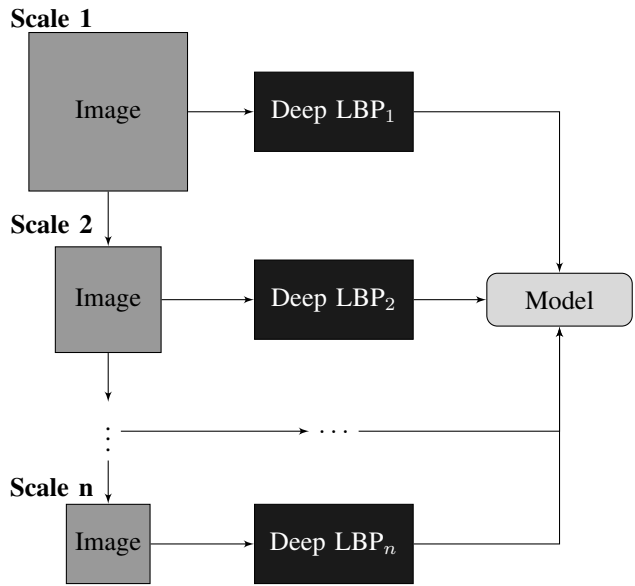
\begin{figure}[!t]
\centering
\begin{tikzpicture}[node distance=3cm, auto]
    \node [image] (img1) {Image};
    \node [image, below of=img1, node distance=2.5cm,
    		   minimum height=4em, minimum width=4em] (img2) {Image};
	\node [below of=img2, node distance = 1.75cm] (imgdots) {$\vdots$};
    \node [image, below of=imgdots, node distance=1.5cm,
    		   minimum height=3em, minimum width=3em] (imgn) {Image};

    \node [deepbox, right of=img1] (lbp1) {Deep LBP$_1$};
    \node [deepbox, right of=img2] (lbp2) {Deep LBP$_2$};
    \node [right of=imgdots] (bboxdots) {$\ldots$};
    \node [deepbox, right of=imgn] (lbpn) {Deep LBP$_n$};

    \node [traditional, right of=lbp2] (model) {Model};

    \path [line] (img1) -- (lbp1);
    \path [line] (img2) -- (lbp2);
    \path [line] (imgn) -- (lbpn);

    \path [line] (img1) -- (img2);
    \path [line] (img2) -- (imgdots);
    \path [line] (imgdots) -- (imgn);

    \path [line] (lbp1) -| (model);
    \path [line] (lbp2) -- (model);
	\path [line] (bboxdots) -| (model);
    \path [line] (lbpn) -| (model);

    \path [line] (imgdots) -- (bboxdots);

	\node at ($(img1) + (-0.75cm, +1.25cm)$) {\textbf{Scale 1}};
	\node at ($(img2) + (-0.75cm, +1cm)$) {\textbf{Scale 2}};
	\node at ($(imgn) + (-0.75cm, +0.75cm)$) {\textbf{Scale n}};
\end{tikzpicture}

\caption{Multi-scale Deep LBP.\label{fig:arch:multiscaledeepLBP}}
\end{figure}

%% file: experiments.tex
\section{Experiments}
\label{sec:experiments}

\input{tables/datasets}

\input{figures/datasets}

In this section, we compare the performance of the proposed
deep LBP architectures against \textit{shallow} LBP versions.
Several datasets were chosen from the LBP literature
covering a wide range of applications, from texture categorization
to object recognition. Table
\ref{tab:datasets} summarizes the datasets used in this work.
Also, Fig. \ref{fig:datasets} shows an image per dataset in order
to understand the task diversity.

We used a 10-fold stratified cross-validation strategy and
the average performance of each method was measured in terms
of:

\begin{itemize}
\item Accuracy.
\item Class rank: Position (\%) of the ground truth label in the ranking
of classes ordered by confidence. The ranking was induced using
probabilistic classifiers.
\end{itemize}

While high values are preferred when using accuracy, low values are
preferred for class rank.
All the images were resized to have a maximum size of $100\times 100$
and the neighborhood used for all the LBP operators was formed
by 8 neighbors on a radius of size 3, which proved to be a good
configuration for the baseline LBP.
The final features were built using a global histogram, without
resorting to image blocks.
Further improvements in each application
can be achieved by fine-tuning the LBP neighborhoods and by using
other spatial sampling techniques on the histogram construction.
Since the objective
of this work was to objectively compare the performance of each
strategy, we decided to fix these parameters.
The final decision model is a Random Forest with 1000 trees.
In the last two datasets, which contain more than 100 classes,
the maximum depth of the decision trees was bounded to 20 in
order to limit the required memory.

In all our experiments, training data was augmented by including vertical and
horizontal flips.


\input{tables/singlescale}

\subsection{Single-scale}

First, we validated the performance of the proposed deep
architectures on single scale settings with increasing
number of deep layers. Information from each layer was
merged at a feature level by concatenating the layerwise
histogram (c.f. Section \ref{sec:architectures:multideeplbp}).
Table \ref{tab:singlescale} summarizes the
results of this setting. In all the datasets, the proposed
models surpassed the results achieved by traditional LBP.

Furthermore, even when the accuracy gains are small, the
large gains in terms of class rank suggest that the deep
architectures induce more stable models, which assign a
high probability on the ground-truth level, even on
misclassified cases. For instance, in the Kylberg dataset,
a small relative accuracy gain of 3.23\% was
achieved by the High Dimensional rule, the relative
gain on the class rank was 69.56\%.

With a few exceptions (e.g. JAFFE dataset), the data-driven
deep operator based on a high dimensional projection
achieved the best performance. Despite the possibility
to induce encoding orderings using user-defined similarity
functions, the final orderings are static and domain independent.
In this sense, more flexible data-driven approaches as the
one suggested in Section \ref{sec:proposal:high} are able
to take advantage of the dataset-specific properties.

Despite the capability of the proposed deep architectures
to achieve large gain margins, 
the deep LBP operators saturate rapidly. For instance,
most of the best results were found on architectures with
up to three deep layers.
Further research
on aggregation techniques to achieve higher levels
of abstraction should be conducted. For instance, it would
be interesting to explore efficient non-parametric approaches
for building encoding orderings that allow more flexible
data-driven optimization.

\subsection{Multi-Scale}

A relevant question in this context is if the observed gains are
due to the higher abstraction levels of the deep LBP encodings
or to the aggregation of information from larger neighborhoods. Namely,
when applying a second order operator, the neighbors of the reference
pixel include information from their own neighborhood which was
initially out of the scope of the local LBP operator.
Thereby, we compare the performance of the Deep LBP and multiscale LBP.

In order to simplify the model assessment, we fixed the number of layers
to 3 in the deep architectures. A scaling factor of 0.5 was used on each
layer of the pyramidal multiscale operator.
Guided by the results achieved in the single-scale experiments, the
deep operator based on the lexicographic sorting of the high-dimensional
feature space was used in all cases.

Table \ref{tab:multiscale} summarizes the results on the multiscale settings.
In most cases, all the deep LBP architectures surpassed the performance of the
best multiscale shallow architecture. Thereby, the aggregation level achieved by
deep LBP operators goes beyond a multiscale analysis, being able to address
meta-texture information. Furthermore, when combined with a multiscale approach,
deep LBP achieved the best results in all the cases.

\FloatBarrier
\input{tables/multiscale}

\subsection{LBPNet}

Finally, we compare the performance of our deep LBP
architecture against the state of the art LBPNet \cite{xi2016local}. As
referred in the introduction, LBPNet uses LBP encodings at
different neighborhood radius and histogram sampling in order
to simulate the process of learning a bag of convolutional
filters in deep networks. Then, the dimensionality of the
descriptors are reduced by means of PCA, resorting to the
idea of pooling layers from Convolutional Neural Networks
(CNN). However, the output of a LBPNet cannot be used by
itself in successive calls of the same function. Thereby, it is
uncapable of building features with higher expresiveness than
the individual operators.

In our experiments, we considered the best LBPNet with
up to three scales and histogram computations with nonoverlapping
histograms that divide the image in $1\times 1$, $2\times 2$
and $3\times 3$ blocks. The number of components kept in the
PCA transformation was chosen in order to retain 95\% of the
variance for most datasets with the exception of 102 Flowers
and Caltech, where a value of 99\% was chosen due to poor
performance of the previous value. A global histogram was
used in our deep LBP architecture

Table \ref{tab:lbpnet} summarizes the results obtained by multiscale
LBP (shallow), LBPNet and our proposed deep LBP. In order to
understand if the gains achieved by the LBPNet are due
to the overcomplete sampling or to the PCA transformation
preceeding the final classifier, we validated the performance
of our deep architecture with a PCA transformation on the
global descriptor before applying the Random Forest classifier.
Despite being able to surpass the performance of our deep
LBP without dimensionality reduction, LBPNet did not improve
the results obtained by our deep architecture with PCA in
most cases. In this sense, even whithout resorting to local
descriptors on the histogram sampling, our model was able
to achieve the best results within the family of LBP methods.
The only exception was observed in the 102 Flowers dataset
(see Fig. \ref{fig:datasets:flowers}), where the spatial
information can be relevant.
It is important to note that our model can also benefit from
using spatial sampling of the LBP activations. Moreover,
deep learning concepts such as dropout and pooling layers
can be introduced within the Deep LBP architectures in a
straightforward manner.

\input{tables/lbpnet}

%% file: tables/datasets.tex
\begin{table}[!t]
\caption{Summary of the datasets used in the
experiments\label{tab:datasets}}
\centering
\begin{tabular}{|c|cccc|}
\hline
\textbf{Dataset} & \textbf{Reference} & \textbf{Task} &
\textbf{Images} & \textbf{Classes}\\
\hline
KTH TIPS & \cite{hayman2004significance} & Texture & 810 & 10\\
FMD & \cite{sharan2009material} & Texture & 1000 & 10\\
Virus & \cite{virusdataset} & Texture & 1500 & 15\\
Brodatz* & \cite{brodatz1966textures} & Texture & 1776 & 111\\
Kylberg & \cite{Kylberg2011c} & Texture & 4480 & 28\\
102 Flowers & \cite{Nilsback08} & Object & 8189 & 102\\
Caltech 101 & \cite{fei2006one} & Object & 9144 & 102\\
\hline
\end{tabular}
\end{table}

%% file: figures/datasets.tex
\def\imgsize{0.18\textwidth}

\begin{figure}[!t]
	\centering
	\begin{subfigure}[t]{\imgsize}
        \centering
        \includegraphics[width=0.9\textwidth,height=0.9\textwidth]
			{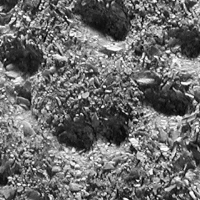}
        \caption{KTH TIPS}
	\end{subfigure}
	\begin{subfigure}[t]{\imgsize}
        \centering
        \includegraphics[width=0.9\textwidth,height=0.9\textwidth]
			{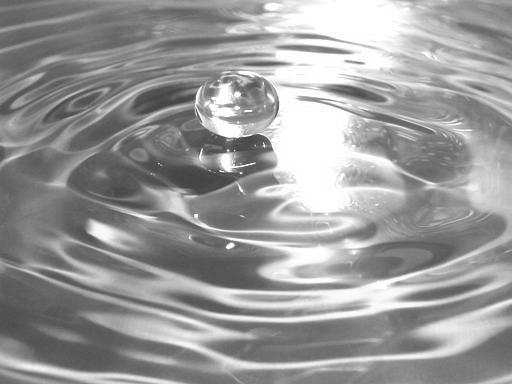}
        \caption{FMD}
	\end{subfigure}

	\begin{subfigure}[t]{\imgsize}
        \centering
        \includegraphics[width=0.9\textwidth,height=0.9\textwidth]
			{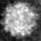}
        \caption{Virus}
	\end{subfigure}
	\begin{subfigure}[t]{\imgsize}
        \centering
        \includegraphics[width=0.9\textwidth,height=0.9\textwidth]
			{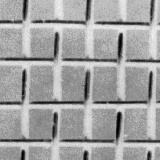}
        \caption{Brodatz}
	\end{subfigure}

	\begin{subfigure}[t]{\imgsize}
        \centering
        \includegraphics[width=0.9\textwidth,height=0.9\textwidth]
			{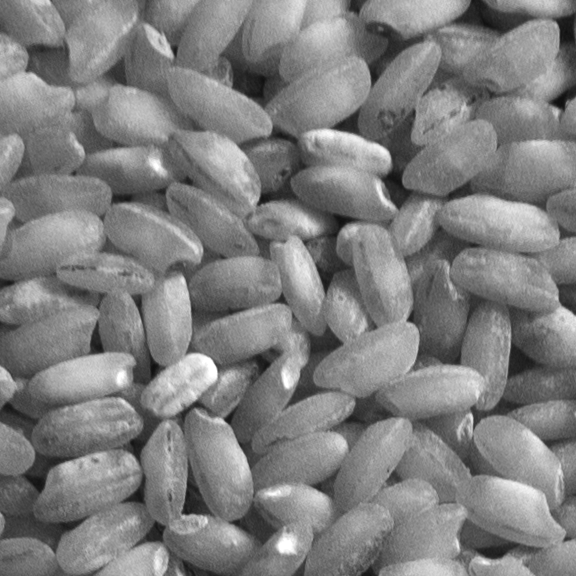}
        \caption{Kylberg}
	\end{subfigure}
	\begin{subfigure}[t]{\imgsize}
        \centering
        \includegraphics[width=0.9\textwidth,height=0.9\textwidth]
			{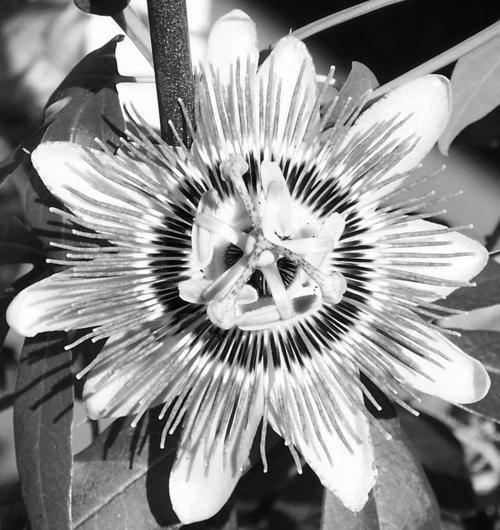}
        \caption{102 Flowers\label{fig:datasets:flowers}}
	\end{subfigure}

	\begin{subfigure}[t]{\imgsize}
        \centering
        \includegraphics[width=\textwidth]
			{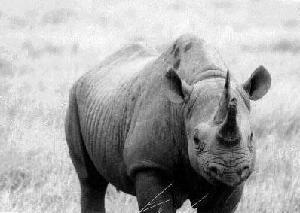}
        \caption{Caltech 101}
	\end{subfigure}

	\caption{Sample images from each dataset
	\label{fig:datasets}}
\end{figure}

%% file: tables/singlescale.tex
\begin{table*}[!t]
\caption{Class rank (\%) of the ground-truth label and accuracy 
with single-scale strategies\label{tab:singlescale}}
\centering
\begin{tabular}{|c|c|ccccc|ccccc|}
\hline
\multirow{3}{*}{\textbf{Dataset}} & \multirow{3}{*}{\textbf{Strategy}} &
\multicolumn{5}{c|}{\textbf{Class Rank}} &
\multicolumn{5}{c|}{\textbf{Accuracy}}\\
\cline{3-12}
& & \multicolumn{5}{c|}{\textbf{Layers}}
& \multicolumn{5}{c|}{\textbf{Layers}}\\
& & \textbf{1} & \textbf{2} & \textbf{3} & \textbf{4} & \textbf{5}
& \textbf{1} & \textbf{2} & \textbf{3} & \textbf{4} & \textbf{5}\\
\hline
&  Similarity & - & 19.18 & 18.87 & \tb{18.26} & 18.80 & - & \tb{53.28} & \tb{53.28} & \tb{53.28} & \tb{53.28}\\
&    High Dim & - & 18.64 & 18.78 & 18.79 & 19.10 & - & \tb{53.28} & \tb{53.28} & \tb{53.28} & \tb{53.28}\\
\hline
\multirow{3}{*}{KTH TIPS}
&         LBP & 1.55 & - & - & - & - & 89.22 & - & - & - & -\\
\cline{2-12}
&  Similarity & - & 1.60 &  1.68 &  1.82 &  1.86 & -     & 88.96 & 88.57 & 86.99 & 87.97\\
&    High Dim & - & 0.94 &  \tb{0.91} &  1.09 &  1.11 & -     & 92.96 & \tb{93.58} & 92.72 & 92.36\\
\hline
\multirow{3}{*}{FMD}
&         LBP & 26.96 & - & - & - & - & 29.20 & - & - & - & -\\
\cline{2-12}
&  Similarity & - & 25.77 & 25.79 & 25.61 & 25.59 & -     &  28.90 & 30.00 & 30.90 & 30.80\\
&    High Dim & - & \tb{23.20} & 23.36 & 23.30 & 23.50 & -     & \tb{33.40} & 32.60 & 33.30 & 33.00\\
\hline
\multirow{3}{*}{Virus}
&         LBP &  8.08 & - & - & - & - & 56.80 & - & - & - & -\\
\cline{2-12}
&  Similarity & - & 6.61 &  6.78 &  6.72 &  6.73  & -     & 61.00 & 61.33 & 60.93 & 61.93\\
&    High Dim & - & 6.65 &  \tb{6.50} &  6.53 &  6.55  & -     & 61.53 & 61.27 & 61.47 & \tb{62.27}\\
\hline
\multirow{3}{*}{Brodatz}
&         LBP &  0.25 & - & - & - & - & 89.23 & - & - & - & -\\
\cline{2-12}
&  Similarity & - & 0.22 &  0.22 &  0.23 &  0.25 & - & 89.73 & 90.23 & 90.50 & 90.36\\
&    High Dim & - & \tb{0.21} &  \tb{0.21} &  0.22 &  0.23 & - & \tb{90.72} & \tb{90.72} & 90.09 & 89.59\\
\hline
\multirow{3}{*}{Kylberg}
&         LBP &  0.23 & - & - & - & - & 95.29 & - & - & - & -\\
\cline{2-12}
&  Similarity & - &  0.18 &  0.16 &  0.14 &  0.14 & - & 96.14 & 96.52 & 96.72 & 96.81\\
&    High Dim & - & \tb{0.07} &  \tb{0.07} &  \tb{0.07} &  \tb{0.07} & - & \tb{98.37} & 98.35 & 98.26 & 98.24\\
\hline
\multirow{3}{*}{102 Flowers}
&         LBP &  13.46 & - & - & - & - & 23.18 & - & - & - & -\\
\cline{2-12}
& Similarity & - & 13.34 & 13.56 & 13.99 & 14.29 & - & \tb{25.59} & 24.46 & 24.92 & 24.58\\
&    High Dim & - & 13.10 & \tb{12.99} & 13.15 & 13.32 & - & 24.56 & 23.76 & 22.81 & 22.36\\
\hline
\multirow{3}{*}{Caltech 101}
&         LBP &  13.05 & - & - & - & - & 39.71 & - & - & - & -\\
\cline{2-12}
& Similarity & - & 12.37 & 12.23 & 12.32 & 12.38 & - & 40.35 & 40.07 & 39.74 & 39.81\\
&    High Dim & - & \tb{11.98} & 12.19 & 12.16 & 12.34 & - & \tb{41.45} & 40.78 & 40.56 & 40.43\\
\hline
\end{tabular}
\end{table*}

%% file: tables/multiscale.tex
\begin{table}[!t]
\caption{Class rank (\%) of the ground-truth label and accuracy 
with multi-scale strategies\label{tab:multiscale}}
\centering
\setlength{\tabcolsep}{0.15cm}
\begin{tabular}{|c|c|ccc|ccc|}
\hline
\multirow{3}{*}{\textbf{Dataset}} & \multirow{3}{*}{\textbf{Strategy}} &
\multicolumn{3}{c|}{\textbf{Class Rank}} &
\multicolumn{3}{c|}{\textbf{Accuracy}}\\
\cline{3-8}
& & \multicolumn{3}{c|}{\textbf{Scales}}
& \multicolumn{3}{c|}{\textbf{Scales}}\\
& & \textbf{1} & \textbf{2} & \textbf{3}
& \textbf{1} & \textbf{2} & \textbf{3}\\
\hline
\hline
\multirow{2}{*}{KTH TIPS}
& Shallow & 1.55 &  1.17 &  1.22 & 89.22 & 90.94 & 90.93\\
&    Deep & 0.91 &  0.79 &  \tb{0.62} & 93.58 & 94.21 & \tb{94.96}\\
\hline
\multirow{2}{*}{FMD}
& Shallow & 26.96 & 26.31 & 26.32 & 29.20 & 29.60 & 29.80\\
&    Deep & \tb{23.36} & 23.54 & 23.77 & 32.60 & \tb{33.20} & 33.00\\
\hline
\multirow{2}{*}{Virus}
& Shallow & 8.08 &  7.51 &  7.97 & 56.80 & 60.60 & 58.60\\
&    Deep & 6.50 &  \tb{5.92} &  6.04 & 61.27 & \tb{66.13} & 64.87\\
\hline
\multirow{2}{*}{Brodatz}
& Shallow & 0.25 &  0.20 &  0.23 & 89.23 & 90.77 & 90.00\\
&    Deep & 0.21 &  \tb{0.13} &  \tb{0.13} & 90.72 & 92.97 & \tb{93.11}\\
\hline
\multirow{2}{*}{Kylberg}
& Shallow & 0.23 &  0.13 &  0.12 & 95.29 & 97.34 & 97.57\\
&    Deep & 0.07 &  0.05 &  \tb{0.04} & 98.35 & 98.84 & \tb{98.95}\\
\hline
\multirow{2}{*}{102 Flowers}
& Shallow & 13.46 & 13.10 & 12.79 & 23.18 & 25.10 & 26.40\\
&    Deep & 12.99 & \tb{12.68} & 12.71 & 23.76 & 26.02 & \tb{26.87}\\
\hline
\multirow{2}{*}{Caltech 101}
& Shallow & 12.92 & 12.46 & 12.28 & 40.07 & 40.84 & 41.03\\
&    Deep & 12.21 & 11.74 & \tb{11.60} & 40.68 & 41.67 & \tb{42.01}\\
\hline
\end{tabular}
\end{table}

%% file: tables/lbpnet.tex
\begin{table}[!t]
\caption{Class rank (\%) of the ground-truth label and accuracy 
\label{tab:lbpnet}}
\centering
\setlength{\tabcolsep}{0.15cm}
\begin{tabular}{|c|c|cc|}
\hline
\textbf{Dataset} & \textbf{Strategy} &
\textbf{Class Rank} & \textbf{Accuracy}\\
\hline
\multirow{4}{*}{KTH TIPS}
& Shallow        & 1.17 & 90.94\\
& LBPNet         & 0.43 & 96.29\\
& Deep LBP       & 0.62 & 94.96\\
& Deep LBP (PCA) & \tb{0.16} & \tb{98.39}\\
\hline
\multirow{4}{*}{FMD}
& Shallow        & 26.31 & 29.80\\
& LBPNet         & 25.71 & 30.00\\
& Deep LBP       & 23.36 & \tb{33.20}\\
& Deep LBP (PCA) & \tb{23.20} & 32.30\\
\hline
\multirow{4}{*}{Virus}
& Shallow        & 7.51 & 60.60\\
& LBPNet         & 7.18 & 60.73\\
& Deep LBP       & 5.92 & \tb{66.13}\\
& Deep LBP (PCA) & \tb{5.91} & 65.60\\
\hline
\multirow{4}{*}{Brodatz}
& Shallow        & 0.20 & 90.77\\
& LBPNet         & 0.20 & 91.49\\
& Deep LBP       & 0.13 & 93.11\\
& Deep LBP (PCA) & \tb{0.12} & \tb{94.46}\\
\hline
\multirow{4}{*}{Kylberg}
& Shallow        & 0.12 & 97.57\\
& LBPNet         & 0.19 & 95.80\\
& Deep LBP       & 0.04 & 98.95\\
& Deep LBP (PCA) & \tb{0.02} & \tb{99.55}\\
\hline
\multirow{4}{*}{102 Flowers}
& Shallow        & 12.79 & 26.40\\
& LBPNet         & \tb{9.61}  & \tb{35.56}\\
& Deep LBP       & 12.68 & 26.87\\
& Deep LBP (PCA) & 22.30 &  8.80\\
\hline
\multirow{4}{*}{Caltech 101}
& Shallow        & 12.46 & 41.03\\
& LBPNet         & 12.11 & 42.69\\
& Deep LBP       & 11.60 & 42.01\\
& Deep LBP (PCA) & \tb{10.87} & \tb{45.14}\\
\hline
\end{tabular}
\end{table}

%% file: conclusions.tex
\section{Conclusions}
\label{sec:conclusions}

Local Binary Patterns have achieved competitive performance in
several computer vision tasks, being a robust and easy to compute
descriptor with high discriminative power on a wide spectrum of tasks.
In this work, we proposed Deep Local Binary Patterns, an extension of
the traditional LBP that allow successive applications of the operator.
By applying LBP in a recursive way, features with higher level of
abstraction are computed that improve the descriptor discriminability.

The key aspect of our proposal is the introduction of flexible
binarization rules that define an order relation between LBP encodings.
This was achieved with two main learning paradigms. First, learning
the ordering based on a user-defined encoding similarity metric. Second,
allowing the user to describe LBP encodings on a high-dimensional space
and learning the ordering on the extended space directly.
Both ideas improved the performance of traditional LBP in a diverse
set of datasets, covering various applications such as face analysis,
texture categorization and object detection. As expected, the paradigm
based on a projection to a high-dimensional space achieved the best
performance, given its capability of using application specific knowledge
in an efficient way. The proposed deep LBP are able to aggregate information
from local neighborhoods into higher abstraction levels, being able to
surpass the performance obtained by multiscale LBP as well.

While the advantages of the proposed approach were demonstrated in the
experimental section, further research can be conducted on several areas.
For instance, it would be interesting to find the minimal properties of interest
that should be guaranteed by the binarization function. In this work, since
we are dealing with intensity-based image, we restricted our analysis to
partial orderings. However, under the presence of other types of data such
as directional (i.e. periodic, angular) data, cycling or local orderings
could be more suitable.
In the most extreme case, the binarization function may be arbitrarily
complex without being restricted to strict orders.

On the other hand, constraining the shape of the binarization function
allows more efficient ways to find suitable candidates. In this sense,
it is relevant to explore ways to improve the performance of the
similarity-based deep LBP. Two possible options would be to refine the
final embedding by using training data and allowing the user to
specify incomplete similarity information.

In this work, each layer was learned in a local fashion, without space
for further refinement. While this idea was commonly used in the
deep learning community when training stacked networks, later
improvements take advantage of refining locally trained
architectures \cite{norouzi2009stacks}. Therefore, we plan to explore global
optimization techniques to refine the layerwise binarization functions.

Deep learning imposed a new era in computer vision and machine learning,
achieving outstanding results on applications where previous
state-of-the-art methods performed poorly.
While the foundations of deep learning rely on
very simple image processing operators, relevant properties
held by traditional methods, such as illumination and rotational
invariance, are not guaranteed.
Moreover, the amount of data required to learn competitive deep models
from scratch is usually prohibitive.
Thereby, it is relevant to explore the path into a unification of
traditional and deep learning concepts. In this work, we explored this
idea within the context of Local Binary Patterns.
The extension of deep concepts to other traditional methods is of
great interest in order to rekindle the
most fundamental concepts of computer vision
to the research community.